# A Novel Design and Improvement of 15-Bar Assembly Tensegrity Robotics Structure.pdf


Yunyi Chu

Advisor: Professor William Keat

Department of Mechanical Engineering

August 2021




**Table of Contents**





# 1. Introduction

A tensegrity, like the one shown in Figure 1, is a flexible structure composed of rigid struts interconnected by pre-tensioned springs. It has a topology that allows it to undergo large deformations when subjected to forces, and then rebound to its original stable shape when those forces are released. When motors are added to the struts (see Fig.2) to impose movement-inducing deformations in a controlled fashion, the result is a unique form of mobile soft-robot. Tensegrity robots can be packed into tight areas and move across a surface by actuating their rigid elements. Tensegrity robots offer a cheaper and less complex alternative to robots with traditional surface methods of movement. However, n-bar tensegrities have some structural limitations. For example, six-bar tensegrities impact orientation sensitivity [1], and they limit the actuation schemes [2].

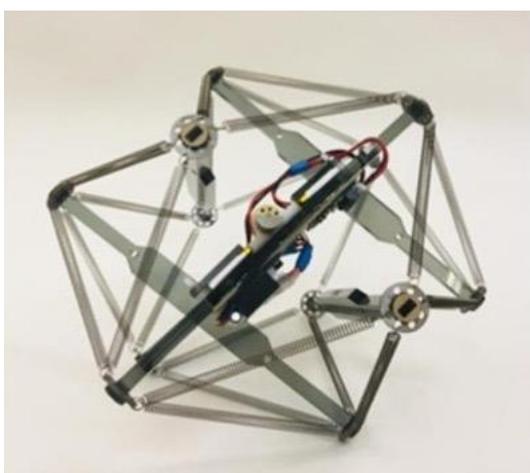

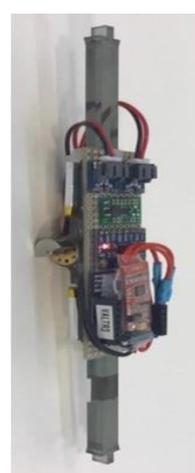

Fig. 1 – Completed wireless 6-Bar tensegrity robot with motors

Fig. 2 – Wireless tensegrity strut module with custom vibration gearmotor and onboard motion capture system

Tensegrities are difficult to assemble because the springs are under tension and the level of difficulty increases with the number of struts. This can limit innovation when designing these soft robots. Therefore, the goal of this summer research was to develop a method for assembling many-barred tensegrities and to demonstrate it by building a model of a 15-bar tensegrity consisting of 15 struts and 78 springs.

Much progress has been made in recent years on the design of these structures. For example,

Li et al. developed a Monte-Carlo-based form finding method [3]. Also, Koohestani [4]



developed an efficient method for finding tensegrity structures using a genetic algorithm. In addition, John Rieffel et al. [5] proposed an evolutionary algorithm that can be used to generate novel large complex tensegrity structures. It offers a robust robotic platform that can rapidly and drastically change shapes by changing string tensions.

To facilitate assembly, Lee et al. fabricated tensegrity structures made of smart materials using 3D printing combined with sacrificial molding. They printed tensegrity with coordinated soft and stiff elements, and designed parameters (such as geometry, topology, density, coordination number, and complexity) which can be used to program system-level mechanics in a soft structure. They endowed the tensegrity with additional functionality by using magnetic materials as tendon components and used a 3D printing technology combined with sacrificial molding to fabricate tensegrities at a diverse scale. This method makes the construction of tensegrity a lot easier because it eliminates any post-assembly process of beam elements.[6]



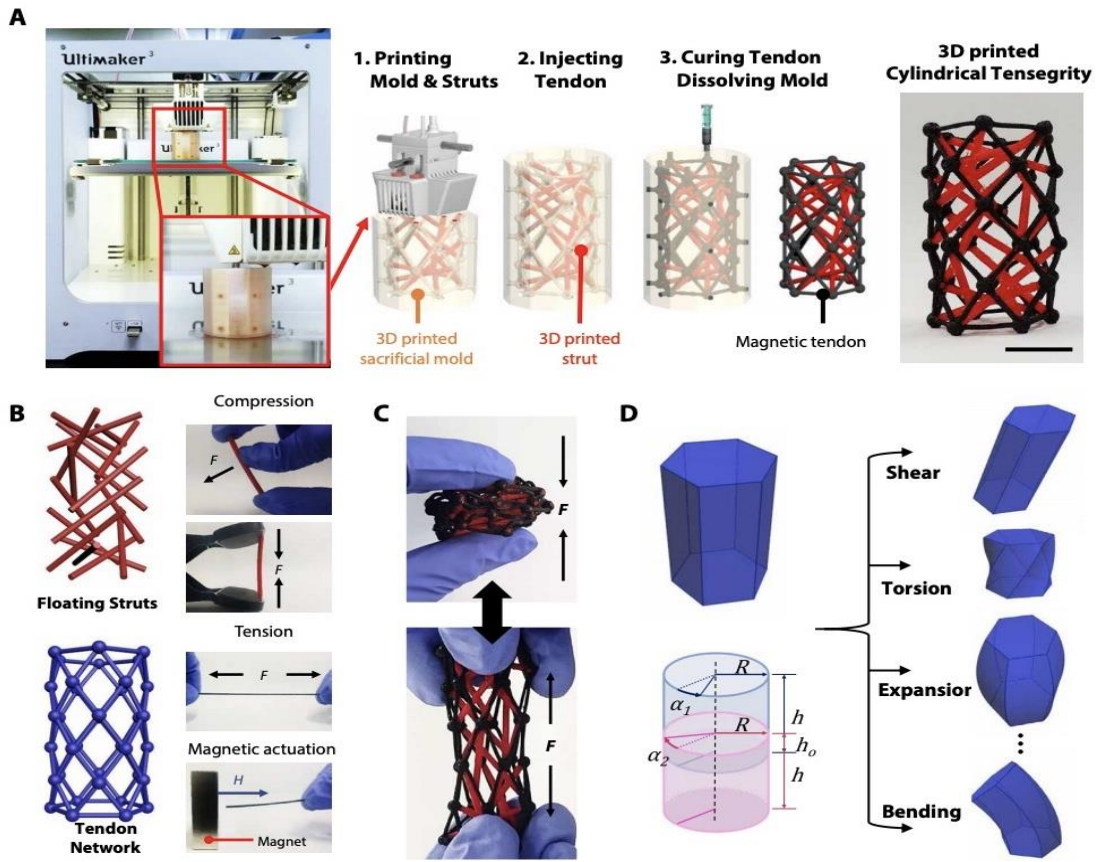

Fig.3- Fabrication Process of the Tensegrity Structure and Mechanical Features of its Element

R. Motro focus on the coupling of tensegrity structures between form and forces, and tensegrity

structural morphology. Structural morphology of tensegrity systems is a key point for tensegrity

systems. He firstly described the design process of the simplest tensegrity system which was

achieved by Kenneth Snelson. Some other simple cells are presented and tensypolyhedra are defined

as tensegrity systems which meet polyhedral geometry in a stable equilibrium state. Secondly, a

numerical model giving access to more complex systems, in terms of number of components and

geometrical properties, is then evoked. His third part is devoted to the linear assemblies of annular

cells can be folded. Some experimental models of the tensegrity ring which is the basic component

of this "hollow rope" have been realized and are examined. In this paper the structural morphology

of tensegrity systems is presented from the simplest cell, the socalled "simplex", to more complex

ones like pentagonal and hexagonal tensegrity rings. The assembly of tensegrity rings provides



interesting structural solutions like the "hollow rope", but one of their main features is their foldability which could be the key for pertinent applications. Other assemblies like woven double layer tensegrity grids can be derived from simple cells, constituting a way from simplicity to complexity. [7] The compressed component created by the chain of 15 struts and its circuit is shown in Figure 4.a., and the physical model of the 15 struts circuit pattern is shown in Figure 4.b.

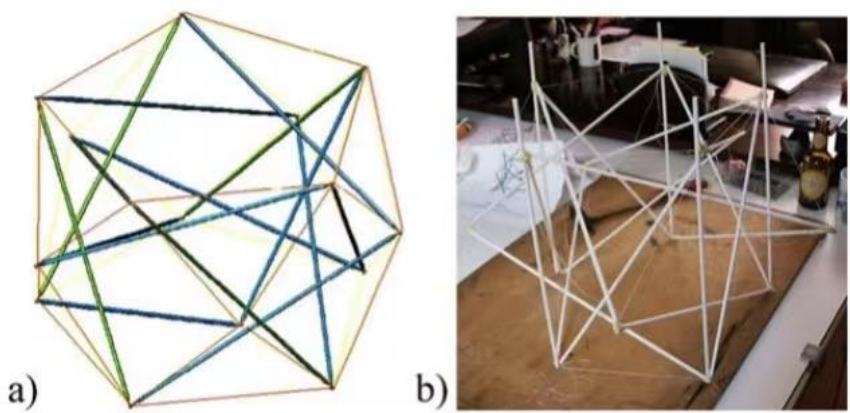

Fig.4- Module assembly (a- Mono circuit tensypolyhedron, b-Physical model)

In contrast with the rapid progress on the design of large n-bar tensegrity structures, the ability to actually implement these designs remains a challenge. Even assembly of a six-bar tensegrity can challenge one's dexterity. Two prior attempts in Prof. Rieffel's lab to assemble a 15-bar tensegrity, e.g., by Chu [8], were unsuccessful. The array of bulky clamps shown in Figure 3, combined with having to deal with a large number of pre-tensioned springs, proved too unwieldy for even the most patient assembler.

Further parts content will focus on design of the 15 bar, computer-assisted assembly of the 15-bar, assembly procedure for a general n-bar tensegrity, and the conclusions of assembly and design of many-bar tensegrity structures according to priority.



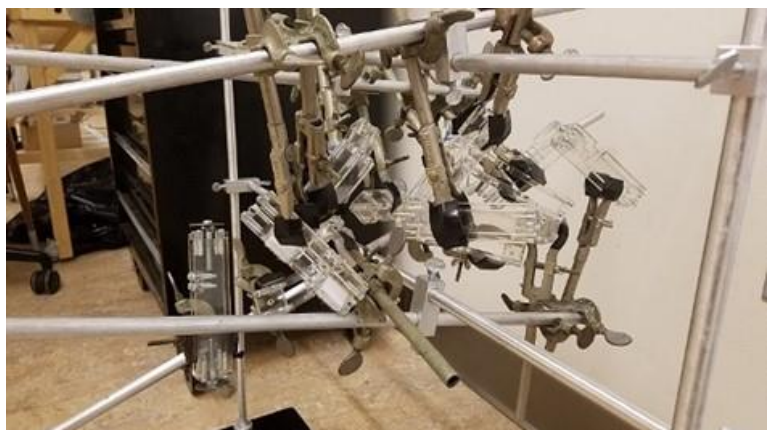

Fig.5- Initial unsuccessful attempt by Chu [8] to assemble
the 15-bar by using a series of clamps to position the struts
based on the results of a form-finding algorithm

## 2. Design of the 15 Bar

Starting point is a computer-generated map of the strut-spring connectivities for a given number of struts. Primary design variables are strut length, spring stiffness, and spring free-length, with the latter having primary control over the shape of the tensegrity. The static equilibrium shape is found by running a form-finding code that minimizes strain energy in the springs. The resulting 15-bar geometry can be seen in Figures 6-8. The suitability of the design is checked by examining if static displacements due to the weights of the struts are acceptable and if natural frequencies are within the operating range of the motors.



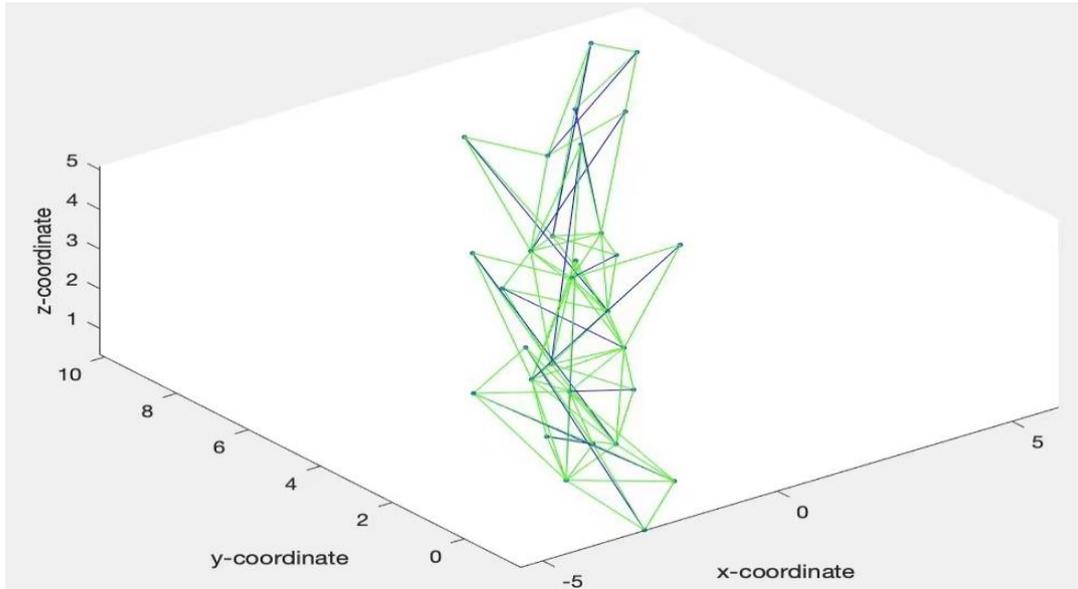

Fig. 6 – Form-Finding 15-Bar by MATLAB

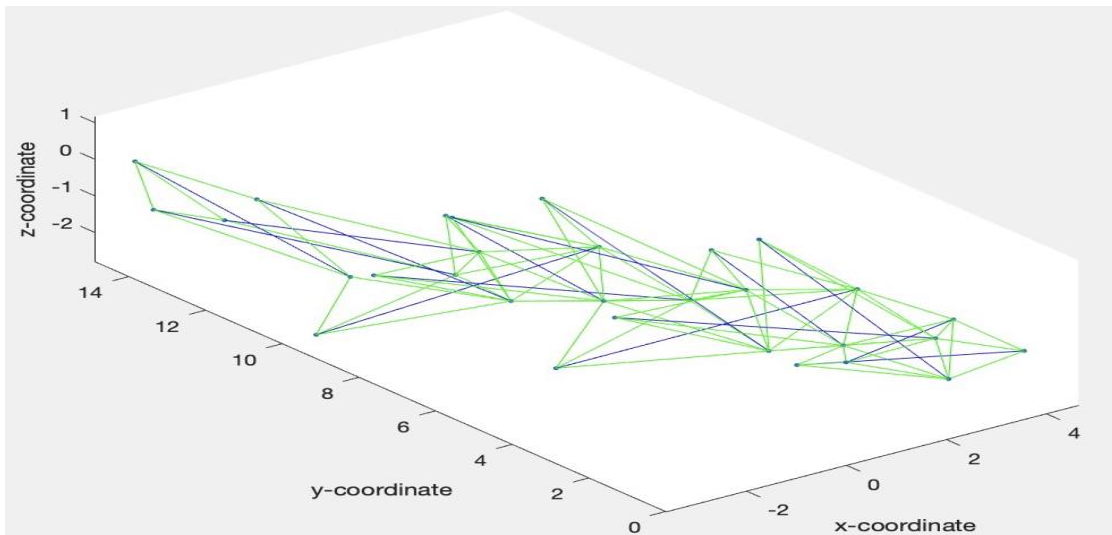

Fig. 7 – Shape After Checking Static Sag

## 3. Computer-Assisted Assembly of the 15 bar

Initial step before assembly is to design and manufacture scaffolding to hold the struts securely in their known positions and orientations so that the springs can be attached. To accomplish this, the 3D computer model of the 15-bar geometry was reoriented so that its longitudinal axis was parallel to the build surface. MATLAB code was then developed to calculate for each strut its centroidal



coordinates and two angles for visually orienting the struts. These variables are defined in Figures 8a and 8b.

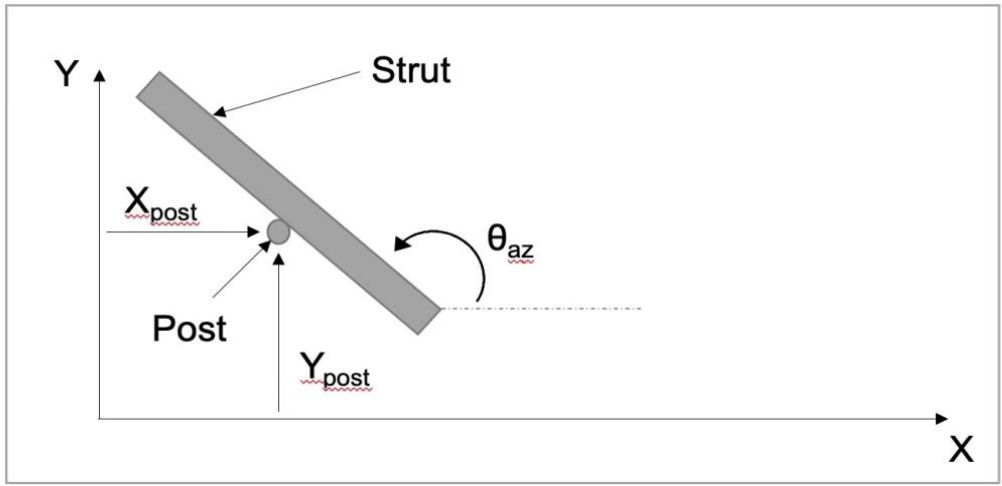

Fig.8a – Top View Scaffoldings Design Variable Definition

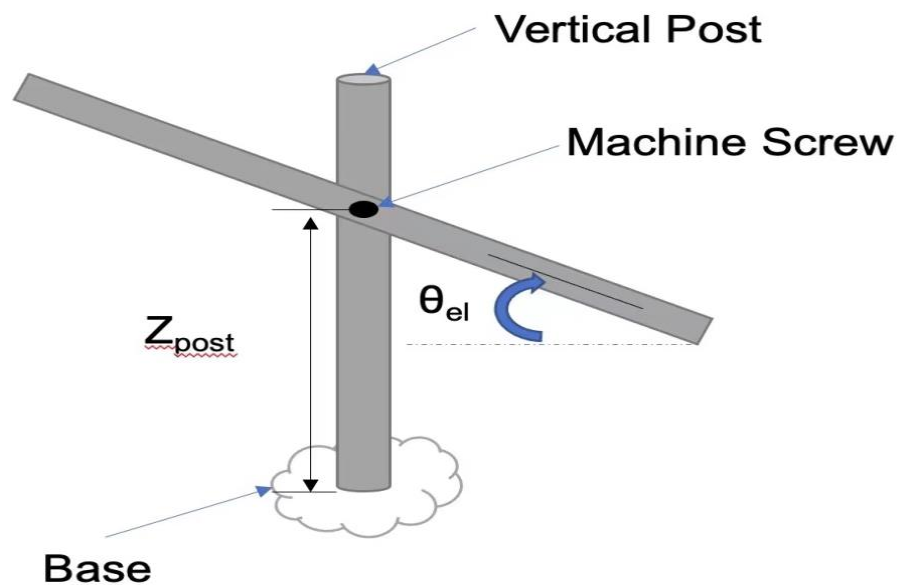

Fig.8b – Front View Scaffoldings Design Variable Definition

The struts were held fixed in space by connecting them to small diameter vertical posts using machine screws and nuts. The spacing of the holes in the plywood base into which the posts were inserted was maximized computationally by searching for the optimal locations of the strut-to-post attachment points. The before and after locations of the holes are shown in Fig.9. The struts and base were then laser-cut from acrylic and plywood respectively. The vertical posts were cut to length from



oak dowels using a hand-saw and the holes for the screws were drilled using a Dremel and drill

press. In Fig.9, the circles are the original positions and the diamonds are the optimized positions.

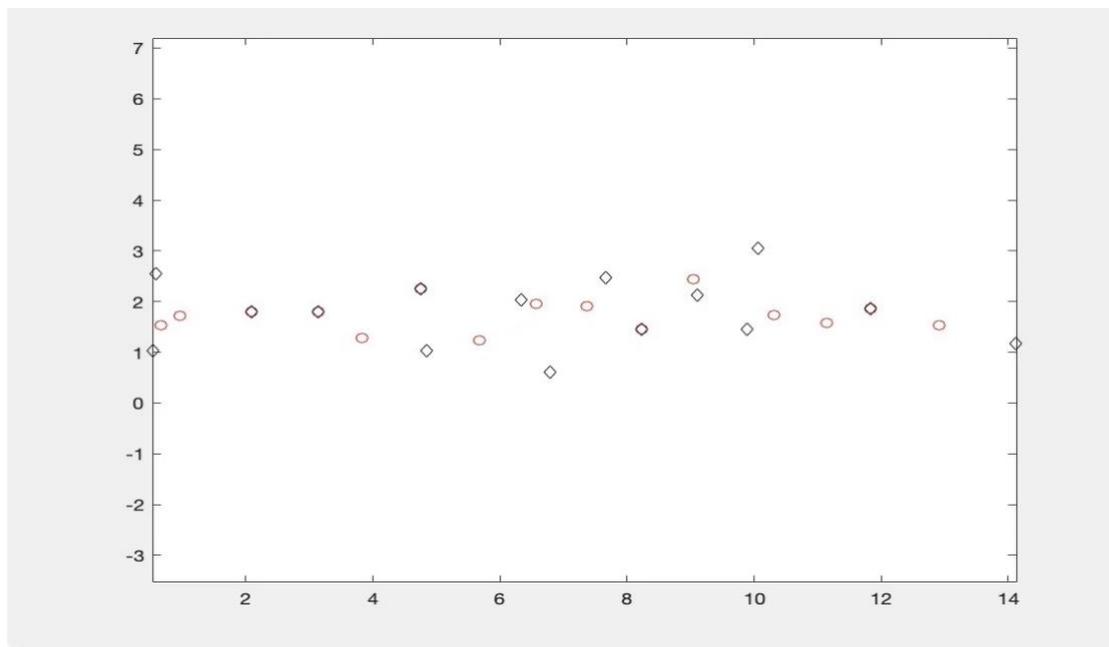

Fig.9 - MATLAB Figure of Base Before and After Shifting of Attachments Points

To assemble the 15-bars, the struts were screwed to the posts and oriented visually based on the

computed azimuth and elevation angles. Last step is to attach 78 helical springs by hooking them

onto perforated washers which have been glued to the ends of the struts. For our model, low cost

elastic cord was used instead of helical springs. Once the screws are removed, the tensegrity can be

lifted off the scaffolding. Fig.10 and 11 show the 15-bar model at different stages of assembly.



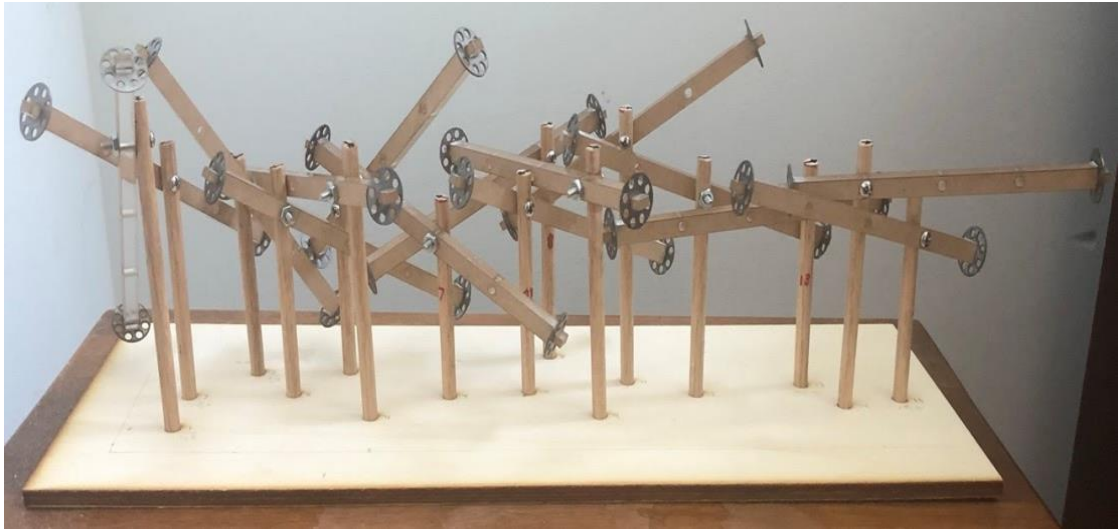

Fig.10 -Scaffolding with Struts Mounted

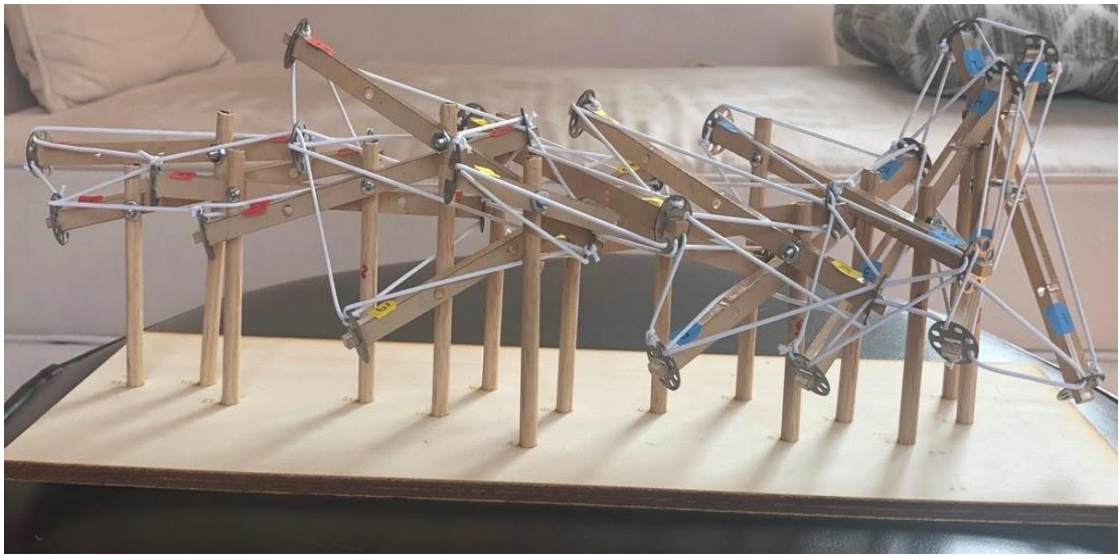

Fig.11 - Full Model of 15-Bar Assembly

## 4. Assembly Procedure for a General n-Bar Tensegrity

DESIGN PROCEDURE

1. **Obtain or generate a topology map** which defines the strut-spring connectivities for the specific n-bar.



2. **Run the form-finding code** (positionFrank.m) to establish the static equilibrium shape of the tensegrity (See Figure 12). The resulting shape can be iterated upon by changing the free length of the springs. Changing the spring stiffness does not change the shape obtained from the form-finding algorithm.

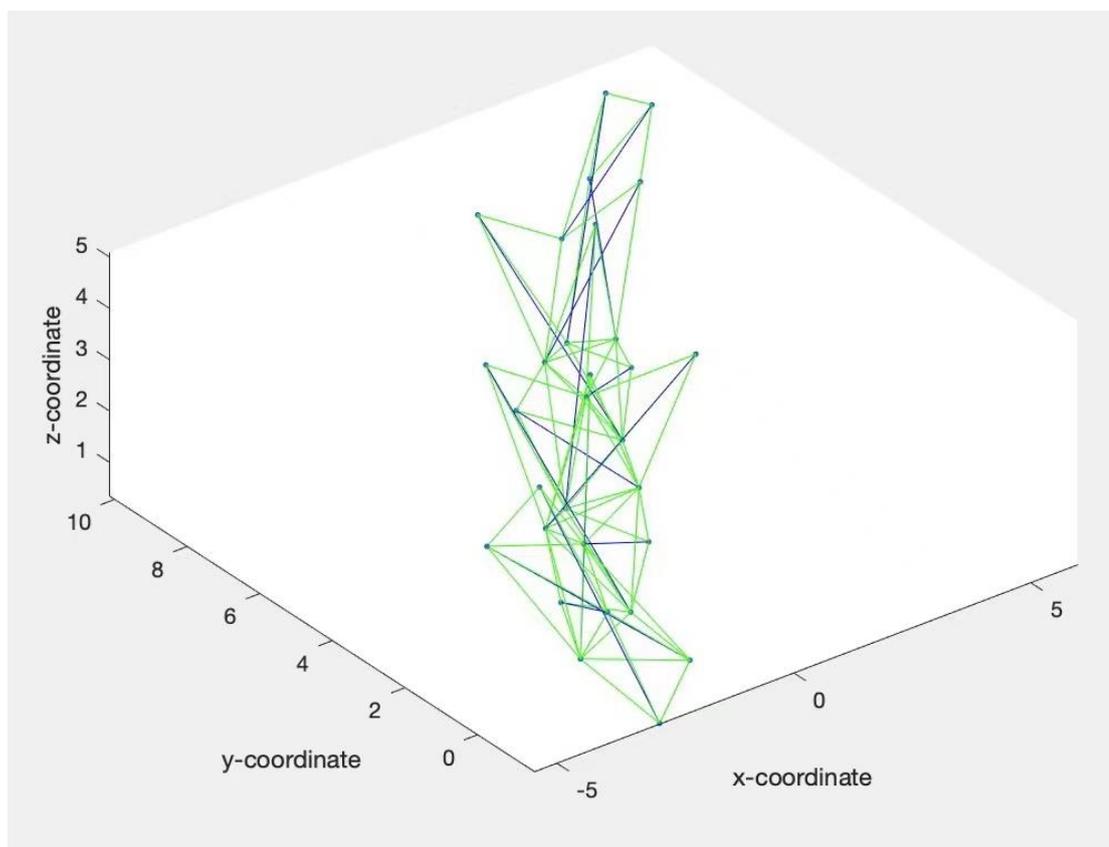

Fig.12- The Static Equilibrium Shape of the Tensegrity

3. **Run additional codes** to check the suitability of the result from the form-finding algorithm.

   o **Stiffa.m** calculates static displacements (i.e. static sag) due to the weights of the struts and also natural frequencies to insure they are within the operating range of the vibration motors. Iterate on spring stiffness to adjust these quantities without changing the shape from the form-finding algorithm,

   o **showmode.m** allows you to animate the mode shapes corresponding to specific natural frequencies.



o **shortdis.m** calculates the shortest distance between each strut and its neighboring struts.

ASSEMBLY PROCEDURE

4. **Design the scaffolding** by running the code **bar15coordinates7.m**. Although named after the 15-bar, this code is applicable to any n-bar tensegrity structure. It performs the following operations:

   o **Calculates the centroids of the struts** in three-dimensional space.

   o **Determines the longitudinal axis**, defined by the two struts whose centroids are furthest apart. It then **reorients the n-bar** in 3D space such that the longitudinal axis is parallel to the global x-axis and also does a global shift to **position the tensegity in the first quadrant of the global coordinates system**.

   o **Plots the n-bar** in its new orientation (See Fig.13).

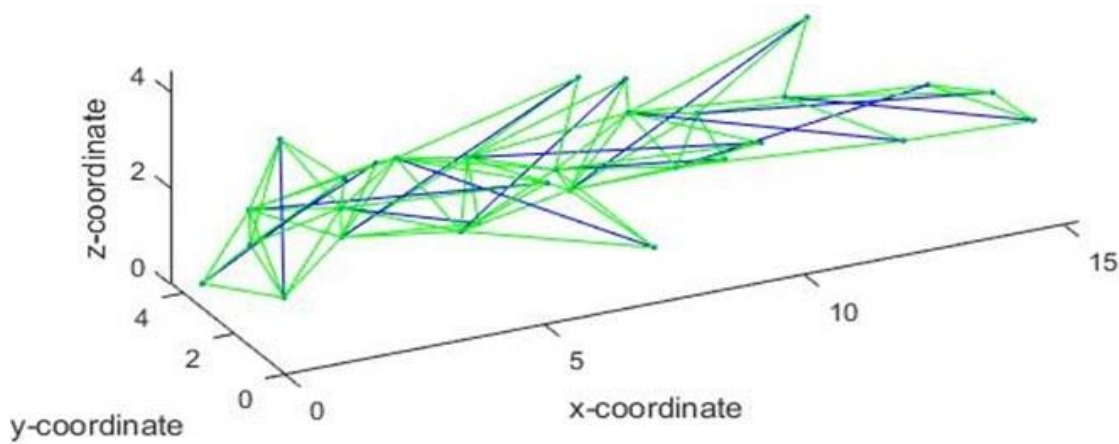

Fig.13- N-Bar In New Orientation After 3D Coordinate Transformation

   o **Calculates the azimuth and elevation angles** for each strut. These angles will be used to visually orient the strut on the vertical posts of the scaffolding. Angles are defined so as to locate the high ends of each strut with respect to the reference lines defined in Figure * and therefore range in value as follows:



$$-180^o \leq \theta_{az} \leq 180^o$$

$$0^o \leq \theta_{el} \leq 90^o$$

- o **Performs an optimization that maximizes the spacing between the vertical posts**. The design variables being optimized are the locations of the strut-post attachment points along each strut. Choices are limited to the ¼, ½ and ¾ points on the struts.

- o **Prints out post coordinates** (xpost, ypost, zpost), **strut angles** (azimuth and elevation), **strut-post attachment positons** (¼, ½ or ¾) and the **z-coordinates of the strut endpoints** so the high side of each strut can be identified (See Appendix A).

5. **Laser-cut the acrylic struts and the plywood base** into which the vertical posts will be inserted. This requires first creating the Solidworks drawings of the strut and base, then submitting the corresponding .dxf files to whoever will be doing the laser-cutting. The Solidworks drawings of the struts included 0.125 in holes at the ¼, ½ or ¾ positions along the strut to provide alternative strut-post attachment points. Strut ends must be designed with 0.25 in by 0.25 in square cross-sections so that perforated steel washers can be slipped on. These washers provide attachment points for the hooked ends of the helical springs.

6. **Create the vertical posts** which will be inserted into the plywood base. For the 15-bar model, these were cut to length from 0.25 in oak dowels using a hand saw. 1.0 in was added to zpost when defining the length of each strut, in order to make allowances for the 0.25 in thick plywood base and to provide 0.5 in of clearance for the lowermost ends of the struts. 0.125 in diameter holes were then drilled 0.25 in down from the top of the posts using a Dremel mounted in a drill press. A center punch was employed to insure that the drill remained centered on the post.



7. **Press fit or glue the vertical posts into the through holes of the plywood base** (See Fig.14). When doing so, the holes in the struts have to be visually aligned using the azimuth angles such that the central axis of the cylindrical hole at the top of the post is perpendicular to the vertical plane in which the strut will lie. For ease of reference, small arrows pointing in the direction defined by the azimuth angle were drawn at the top of each vertical post using a Sharpie.

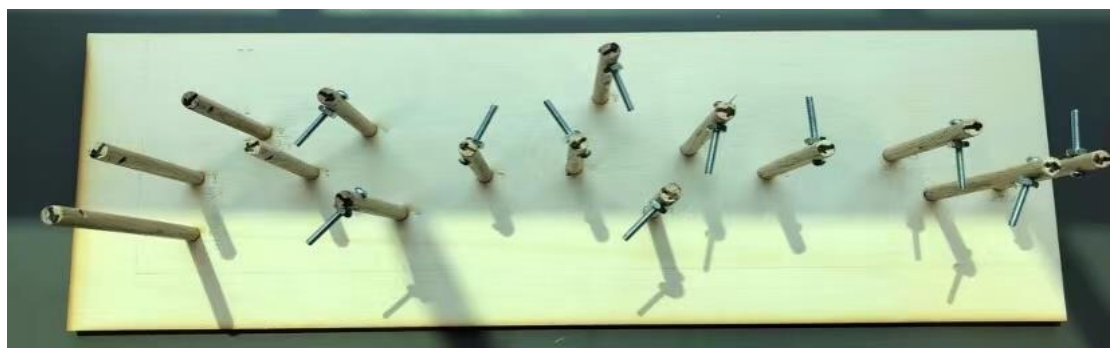

Fig.14- The Plywood Base with the Vertical Posts

8. **Superglue the perforated washers to the strut ends.**

9. **Mount the struts to the vertical posts** using ¾ in long #4 machine screws and nuts. Before tightening the nuts, care must be taken to orient each strut in accordance with its elevation angle (See Fig.15).

10. **Attach the springs to the struts** by hooking the ends of the helical springs[1] onto the perforated washers in a manner consistent with the topology map of step 1. To facilitate identifying where the springs should be attached, each strut endpoint was labeled with a sticker indicating its vertex number (See Fig. 16).

---

[1] For the model, elastic cord was used instead of helical springs



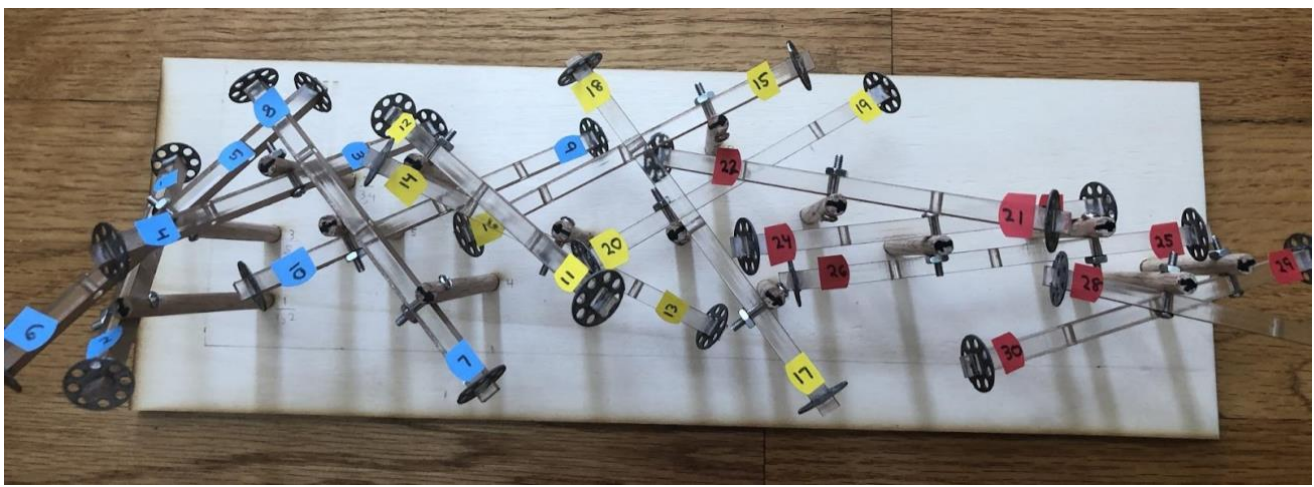

Fig. 15- 15-Bar Assembly with The Struts Mounted to The Vertical Posts

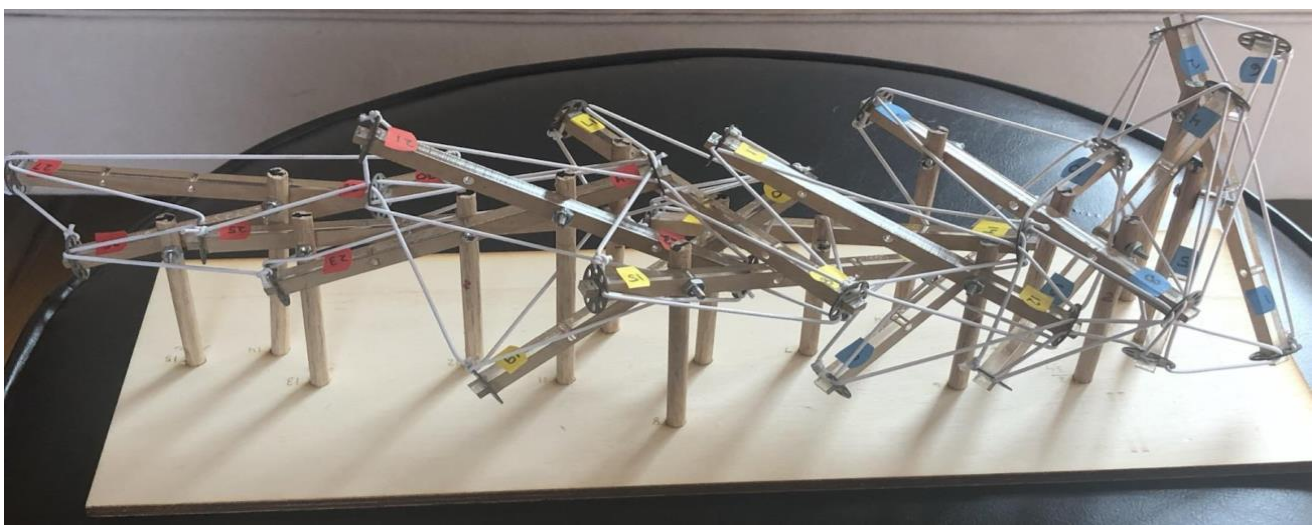

Fig. 16- 15-Bar Assembly with Elastic Cord (In Place of Springs) Attached to The Struts

## 5. Conclusion

The assembly method was effective at creating the model of the 15-bar tensegrity and is applicable to other tensegrity structures. Based on our experience, we estimate that the scaffolding can be designed and manufactured in 1-2 days and that full assembly can be achieved in 1 day, once the assembler is familiar with the process.



The key to further simplifying the assembly process may lie with improving the design of the tensegrity itself. For example, the computational design tools could be recast in the form of an optimization code to maximize spacing between struts and springs.

**Appendix A – Values Printed Out by bar15coordinates7.m.**

pt1 =

     1

pt2 =

    15

maxdis =

   12.2436

xpost =

  Columns 1 through 8

| 0.5302 | 2.0914 | 0.5910 | 3.1395 | 4.8495 | 4.7545 | 6.7826 | 7.6641 |

  Columns 9 through 15

| 6.3257 | 8.2338 | 10.0656 | 9.0981 | 9.8843 | 11.8290 | 14.1256 |

ypost =

  Columns 1 through 8

| 1.0297 | 1.7995 | 2.5582 | 1.7918 | 1.0261 | 2.2556 | 0.6140 | 2.4785 |

  Columns 9 through 15

| 2.0287 | 1.4548 | 3.0474 | 2.1279 | 1.4594 | 1.8611 | 1.1660 |

zpost =

  Columns 1 through 8

| 1.4161 | 1.4465 | 3.2302 | 2.5618 | 0.7701 | 1.9557 | 2.3657 | 1.0290 |



Columns 9 through 15

  1.8610      1.6091      2.5676      2.3247      2.1276      2.3783      2.3499

thetael =

  Columns 1 through 8

  79.9774     46.7932     58.0070     45.1947     48.0388     37.7851      6.3675     26.1570

  Columns 9 through 15

  46.5061     42.0398     27.4581      3.2837      5.9276     27.8341     12.3356

thetaaz =

  Columns 1 through 8

  74.4751  -168.9037    114.0598     -1.6195    165.7398     32.6991    -29.1135  -154.4588

  Columns 9 through 15

  -6.2954    170.7188     31.0561    -17.6704  -174.4903     -7.2029    163.3519

jsave =

  Columns 1 through 14

    1      2      3      2      1      2      3      1      1      2      3      1      3
2

  Column 15

    1